\pdfoutput=1

\documentclass[11pt]{article}

\usepackage{acl}

\usepackage{times}
\usepackage{latexsym}

\usepackage[T1]{fontenc}

\usepackage[utf8]{inputenc}

\usepackage{microtype}

\usepackage{subfigure}
\usepackage{graphicx}
\usepackage{algorithm}
\usepackage{algorithmic}
\usepackage{multirow}
\usepackage{booktabs}
\usepackage{amssymb}
\usepackage{subfigure}	
\usepackage{bm}
\mathchardef\mhyphen="2D
\usepackage{amsmath}
\usepackage{makecell}
\usepackage{color}
\title{Learning Confidence for Transformer-based Neural Machine Translation}

\author{Yu Lu\textsuperscript{1,2}\thanks{\ \ Work done while the author was an intern at Tencent.}, Jiali Zeng\textsuperscript{3}, Jiajun Zhang\textsuperscript{1,2}\thanks{\ \ Corresponding author.}, Shuangzhi Wu\textsuperscript{3} and Mu Li\textsuperscript{3} \\
\textsuperscript{1} National Laboratory of Pattern Recognition, Institute of Automation, CAS, Beijing, China \\
\textsuperscript{2} School of Artificial Intelligence, University of Chinese Academy of Sciences, Beijing, China \\
\textsuperscript{3} Tencent Cloud Xiaowei, Beijing, China\\
\texttt{$\left\{\right.$yu.lu, jjzhang$\left .\right\}$@nlpr.ia.ac.cn} \\
\texttt{$\left\{\right.$lemonzeng, frostwu, ethanlli$\left .\right\}$@tencent.com}
}

\begin{document}
\maketitle
\begin{abstract}
Confidence estimation aims to quantify the confidence of the model prediction, providing an expectation of success. A well-calibrated confidence estimate enables accurate failure prediction and proper risk measurement when given noisy samples and out-of-distribution data in real-world settings. However, this task remains a severe challenge for neural machine translation (NMT), where probabilities from softmax distribution fail to describe when the model is probably mistaken. To address this problem, we propose an unsupervised confidence estimate learning jointly with the training of the NMT model. We explain confidence as how many hints the NMT model needs to make a correct prediction, and more hints indicate low confidence. Specifically, the NMT model is given the option to ask for hints to improve translation accuracy at the cost of some slight penalty. Then, we approximate their level of confidence by counting the number of hints the model uses. We demonstrate that our learned confidence estimate achieves high accuracy on extensive sentence/word-level quality estimation tasks. Analytical results verify that our confidence estimate can correctly assess underlying risk in two real-world scenarios: (1) discovering noisy samples and (2) detecting out-of-domain data. We further propose a novel confidence-based instance-specific label smoothing approach based on our learned confidence estimate, which outperforms standard label smoothing\footnote{https://github.com/yulu-dada/Learned-conf-NMT}.
\end{abstract}

\section{Introduction}

Confidence estimation has become increasingly critical with the widespread deployment of deep neural networks in practice \citep{DBLP:journals/corr/AmodeiOSCSM16}. It aims to measure the model's confidence in the prediction, showing when it probably fails. A calibrated confidence estimate can accurately identify failure, further measuring the potential risk induced by noisy samples and out-of-distribution data prevalent in real scenarios \citep{nguyen2015posterior,DBLP:conf/nips/SnoekOFLNSDRN19}.

\begin{figure}[t]
	\centering
	\includegraphics[width=.475\textwidth]{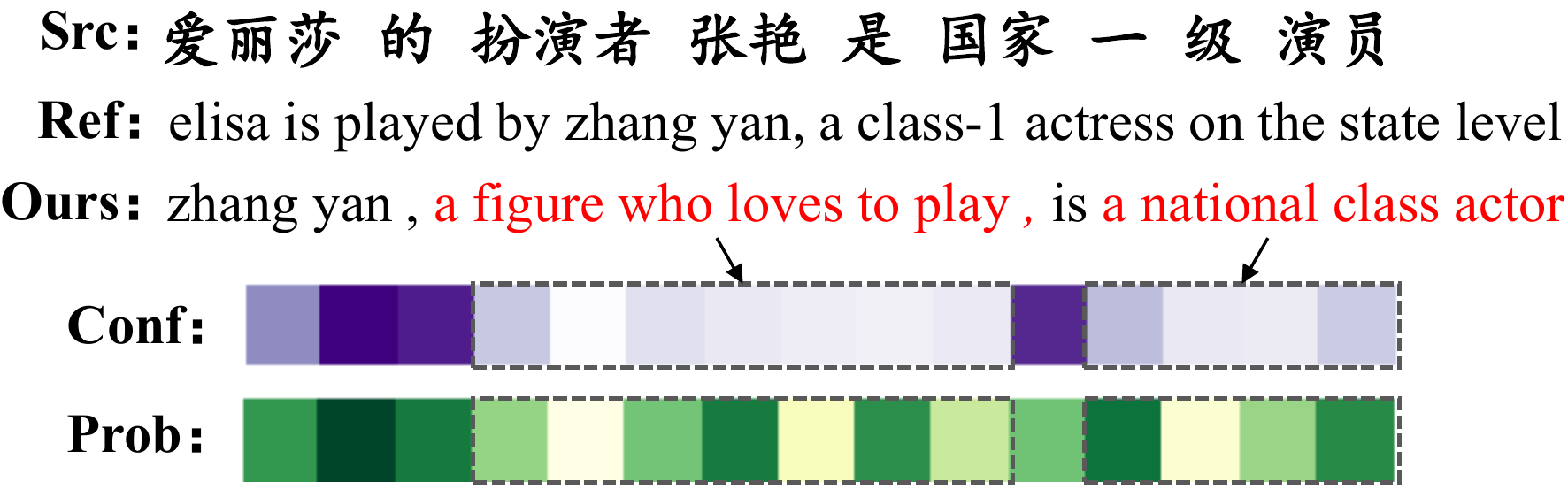}
	\caption{An example of generated probabilities and our learned confidence estimates. The phrases in red are wrong translations. The corresponding prediction probabilities and confidence estimates are outlined in dashed boxes. The dark color indicates a large value under two evaluations.}
	\label{example}
\end{figure}

Unfortunately, neural machine translation (NMT) is reported to yield poor-calibrated confidence estimate \citep{DBLP:journals/corr/abs-1903-00802,DBLP:conf/acl/WangTSL20}, which is common in the application of modern neural networks \citep{DBLP:conf/icml/GuoPSW17}. It implies that the probability a model assigns to a prediction is not reflective of its correctness. Even worse, the model often fails silently by providing high-probability predictions while being woefully mistaken \citep{DBLP:conf/iclr/HendrycksG17}. We take Figure~\ref{example} as an example. The mistranslations are produced with high probabilities (dark green blocks in the dashed box), making it problematic to assess the quality based on prediction probability when having no access to references.

The confidence estimation on classification tasks is well-studied in the literature \citep{platt1999probabilistic,DBLP:conf/icml/GuoPSW17}. Yet, researches on structured generation tasks like NMT is scarce. Existing researches only study the phenomenon that the generated probability in NMT cannot reflect the accuracy \citep{DBLP:conf/nips/MullerKH19, DBLP:conf/acl/WangTSL20}, while little is known about how to establish a well-calibrated confidence estimate to describe the predictive uncertainty of the NMT model accurately.

To deal with this issue, we aim to learn the confidence estimate jointly with the training process in an unsupervised manner. Inspired by \textit{Ask For Hints} \citep{DBLP:journals/corr/abs-1802-04865}, we explain confidence as how many hints the NMT model needs to make a correct prediction. Specifically, we design a scenario where ground truth is available for the NMT model as hints to deal with tricky translations. But each hint is given at the price of some penalty. Under this setting, the NMT model is encouraged to translate independently in most cases to avoid penalties but ask for hints to ensure a loss reduction when uncertain about the decision. More hints mean low confidence and vice versa. In practice, we design a confidence network, taking multi-layer hidden states of the decoder as inputs to predict the confidence estimate. Based on this, we further propose a novel confidence-based label smoothing approach, in which the translation more challenging to predict has more smoothing to its labels.

Recall the example in Figure~\ref{example}. The first phrase ``\textit{a figure who loves to play}'' is incorrect, resulting in a low confidence level under our estimation. We notice that the NMT model is also uncertain about the second expression ``\textit{a national class actor}'', which is semantically related but has inaccurate wording. The translation accuracy largely agrees with our learned confidence rather than model probabilities.

We verify our confidence estimate as a well-calibrated metric on extensive sentence/word-level quality estimation tasks, which is proven to be more representative in predicting translation accuracy than existing unsupervised metrics \citep{DBLP:FomichevaSYBGFA20}. Further analyses confirm that our confidence estimate can precisely detect potential risk caused by the distributional shift in two real-world settings: separating noisy samples and identifying out-of-domain data. The model needs more hints to predict fake or tricky translations in these cases, thus assigning them low confidence. Additionally, experimental results show the superiority of our confidence-based label smoothing over the standard label smoothing technique on different-scale translation tasks (WMT14 En$\Rightarrow$De, NIST Zh$\Rightarrow$En, WMT16 Ro$\Rightarrow$En, and IWSLT14 De$\Rightarrow$En).


The contributions of this paper are three-fold:
\begin{itemize}
    \item We propose the learned confidence estimate to predict the confidence of the NMT output, which is simple to implement without any degradation on the translation performance.
    \item We prove our learned confidence estimate as a better indicator of translation accuracy on sentence/word-level quality estimation tasks. Furthermore, it enables precise assessment of risk when given noisy data with varying noise degrees and diverse out-of-domain datasets. 
    \item We design a novel confidence-based label smoothing method to adaptively tune the mass of smoothing based on the learned confidence level, which is experimentally proven to surpass the standard label smoothing technique.
\end{itemize}

\section{Background}

In this section, we first briefly introduce a mainstream NMT framework, Transformer \citep{vaswani2017attention}, with a focus on how to generate prediction probabilities. Then we present an analysis of the confidence miscalibration observed in NMT, which motivates our ideas discussed afterward.

\subsection{Transformer-based NMT}

The Transformer has a stacked encoder-decoder structure. When given a pair of parallel sentences ${x}=\left\lbrace x_1,x_2,...x_S \right\rbrace$ and ${y}=\left\lbrace y_1,y_2,...y_T \right\rbrace$, the encoder first transforms input to a sequence of continuous representations ${\bm h}=\left\lbrace \bm{h^0_1,h^0_2,...h^0_S}\right\rbrace$, which are then passed to the decoder.

The decoder is composed of a stack of $N$ identical blocks, each of which includes self-attention, cross-lingual attention, and a fully connected feed-forward network. The outputs of $l$-th block $h^l_t$ are fed to the successive block. At the $t$-th position, the model produces the translation probabilities $p_t$, a vocabulary-sized vector, based on outputs of the $N$-th layer:
\begin{equation}
    p_t={\rm softmax}(\bm{W}\bm{h^N_t}+b)
\end{equation}

During training, the model is optimized by minimizing the cross entropy loss:
\begin{equation}
    \mathcal{L_{\rm NMT}}=\sum_{t=1}^{T}-y_t{\rm log}(p_t)
\end{equation}
where $\left\{\bm{W}, b\right\}$ are trainable parameters and $y_t$ is denoted as a one-hot vector. During inference, we implement beam search by selecting high-probability tokens from generated probability for each step. 

\subsection{Confidence Miscalibration in NMT}

Modern neural networks have been found to yield a miscalibrated confidence estimate \citep{DBLP:conf/icml/GuoPSW17,DBLP:conf/iclr/HendrycksG17}. It means that the prediction probability, as used at each inference step, is not reflective of its accuracy. The problem is more complex for structured outputs in NMT. We cannot judge a translation as an error, even if it differs from the ground truth, as several semantically equivalent translations exist for the same source sentence. Thus we manually annotate each target word as OK or BAD on 200 Zh$\Rightarrow$En translations. Only definite mistakes are labeled as BAD, while other uncertain translations are overlooked.

\begin{figure}
	\centering
	\includegraphics[width=.49\textwidth]{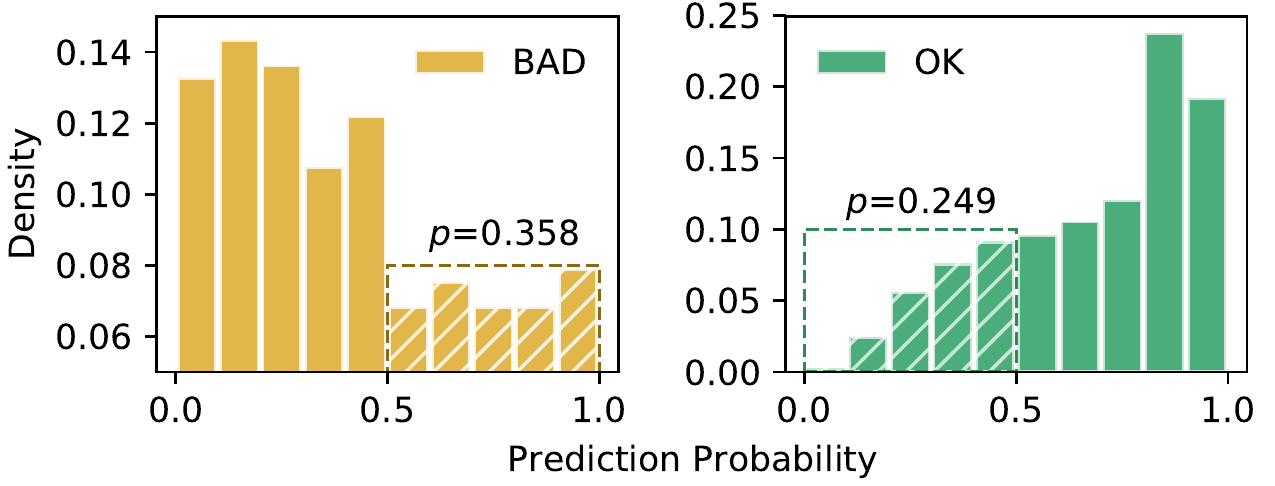}
	\caption{The density function of word probabilities predicted by the NMT model on OK and BAD translations. We outline the miscalibration with slash mark: \textit{over-confident} (producing high probabilities for errors) and \textit{under-confident} (generating low probabilities for right translations).}
	\label{fig:mis}
\end{figure}

Figure~\ref{fig:mis} reports the density function of prediction probabilities on OK and BAD translations. We observe severe miscalibration in NMT: \textit{over-confident} problems account for 35.8\% when the model outputs BAD translations, and 24.9\% OK translations are produced with low probabilities. These issues make it challenging to identify model failure. It further drives us to establish an estimate to describe model confidence better. 

\section{Learning to Estimate Confidence}

\label{method}

\begin{figure*}[t]
	\centering
	\includegraphics[width=.85\textwidth]{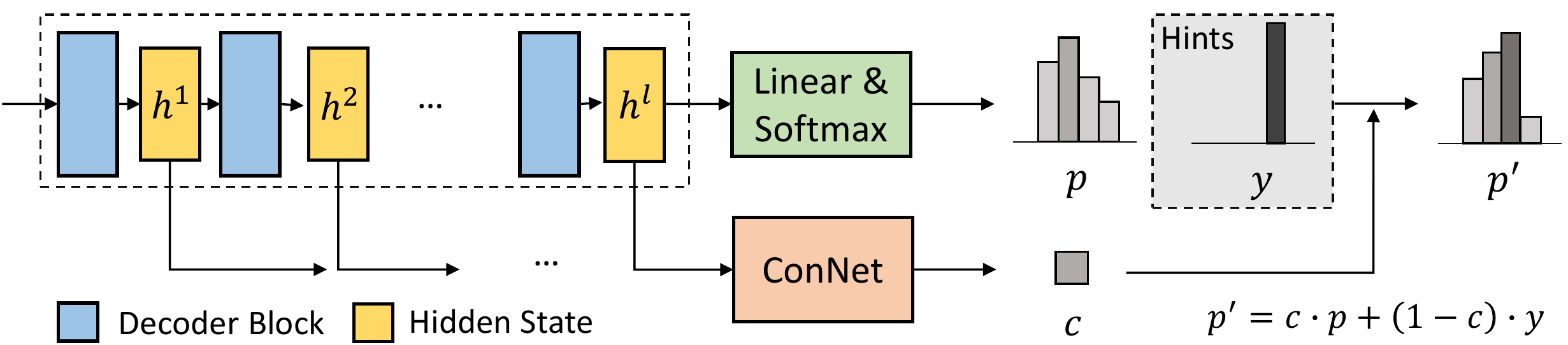}
	\caption{The overview of the framework. The NMT model is allowed to ask for hints (ground-truth translation) during training based on the confidence level predicted by the ConNet. During inference, we use the model prediction $p$ to sample hypotheses. Each translation word comes with a corresponding confidence estimate.}
	\label{fig:framework}
\end{figure*}

A well-calibrated confidence estimate should be able to tell when the NMT model probably fails. Ideally, we would like to learn a measure of confidence for each target-side translation, but this remains a thorny problem in the absence of ground truth for confidence estimate. Inspired by \textit{Ask For Hints} \cite{DBLP:journals/corr/abs-1802-04865} on the image classification task, we define confidence as how many hints the NMT model needs to produce the correct translation. More hints mean low confidence, and that is a high possibility of failure.

\paragraph{Motivation.} We assume that the NMT model can ask for hints (look at ground-truth labels) during training, but each clue comes at the cost of a slight penalty. Intuitively, a good strategy is to independently make the predictions that the model is confident about and then ask for clues when the model is uncertain about the decision. Under this assumption, we approximate the confidence level of each translation by counting the number of hints used.  

To enable the NMT model to ask for hints, we add a confidence estimation network (ConNet) in parallel with the original prediction branch, as shown in Figure~\ref{fig:framework}. The ConNet takes hidden states of the decoder at $t$-th step ($\bm h_t$) as inputs and predicts a single scalar between 0 and 1. 
\begin{equation}
    c_t=\sigma(\bm{W^{'}}\bm{h_t}+b^{'})
\end{equation}
where $\theta_c=\{W^{'},b^{'}\}$ are trainable parameters. $\sigma(\cdot)$ is the sigmoid function. If the model is confident that it can translate correctly, it should output $c_t$ close to 1. Conversely, the model should output $c_t$ close to 0 for more hints.

To offer the model ``hints'' during training, we adjust softmax prediction probabilities by interpolating the ground truth probability distribution $y_t$ (denoted as a one-hot vector) into the original prediction. The degree of interpolation is decided by the generated confidence $c_t$:
\begin{equation}
    p_t^{'}=c_t\cdot{p_t}+(1-c_t)\cdot{y_t}
    \label{eqa:2}
\end{equation}
The translation loss is calculated using modified prediction probabilities. 
\begin{equation}
    \mathcal{L_{\rm NMT}}=\sum_{t=1}^{T}-y_t{\rm log}(p_t^{'})
\end{equation}

To prevent the model from minimizing the loss by always setting $c_t=0$ (receiving all the ground truth), we add a $\rm log$ penalty to the loss function. 
\begin{equation}
    \mathcal{L_{\rm Conf}}=\sum_{t=1}^{T}-{\rm log}(c_t)
\end{equation}
The final loss is the sum of the translation loss and the confidence loss, which is weighted by the hyper-parameter $\lambda$:
\begin{equation}
    \mathcal{L}=\mathcal{L_{\rm NMT}}+\lambda\mathcal{L_{\rm Conf}}
    \label{loss}
\end{equation}

Under this setting, when $c\to 1$ (the model is quite confident), we can see that $p^{'}\to p$ and $\mathcal{L_{\rm Conf}}\to 0$, which is equal to a standard training procedure. In the case where $c\to 0$ (the model is quite unconfident), we see that $p^{'}\to y$ (the model obtains correct labels). In this scenario, $\mathcal{L_{\rm NMT}}$ would approach 0, but $\mathcal{L_{\rm Conf}}$ becomes very large. Thus, the model can reduce the overall loss only when it successfully predicts which outputs are likely to be correct.

\paragraph{Implementation Details.}

Due to the complexity of Transformer architecture, it requires several optimizations to prevent the confidence branch from degrading the performance of the translation branch.

\textit{Do not provide hints at the initial stage.} The early model is fragile, which lays the groundwork for the following optimization. We find that affording hints at an early period leads to a significant performance drop. To this end, we propose to dynamically control the value of $\lambda$ (as in Equation~\ref{loss}) by the training step ($s$) as:
\begin{equation}
    \lambda(s)=\lambda_0*e^{-s/{\beta_0}}
    \label{anneal}
\end{equation}
where $\lambda_0$ and $\beta_0$ control the initial value and the declining speed of $\lambda$. We expect the weight of confidence loss to be large at the beginning ($c\rightarrow 1$) and give hints during middle and later stages.

\textit{Do not use high-layer hidden states to predict confidence.} We find that it would add much burden to the highest layer hidden state if used to predict translation and confidence simultaneously. So we suggest using low-layer hidden states for the confidence branch and leaving the translation branch unchanged (here, the decoder has 6 layers):
\begin{equation}
    {\bm h_t}={\rm AVE}({\bm h_t^1}+{\bm h_t^2}+{\bm h_t^3})
\end{equation}
where $h_t^l$ is the $l$-th layer hidden state in the decoder. Besides, other combinations of low-layer hidden states are alternative, i.e., ${\bm h_t}={\rm AVE}({\bm h_t^1}+{\bm h_t^3})$.

\textit{Do not let the model lazily learn complex examples.} We encounter the situation where the model frequently requests hints rather than learning from difficulty. We follow \citet{DBLP:journals/corr/abs-1802-04865} to give hints with 50\% probability. In practice, we apply Equation~\ref{eqa:2} to only half of the batch.

\paragraph{Confidence-based Label Smoothing.} Smoothing labels is a typical way to prevent the network from miscalibration \citep{DBLP:conf/nips/MullerKH19}. It has been used in many state-of-the-art models, which assigns a certain probability mass ($\epsilon_0$) to other non-ground-truth labels \citep{DBLP:conf/cvpr/SzegedyVISW16}. Here we attempt to employ our confidence estimate to improve smoothing. We propose a novel instance-specific confidence-based label smoothing technique, where predictions with greater confidence receive less label smoothing and vice versa. The amount of label smoothing applied to a prediction ($\epsilon_{t}$) is proportional to its confidence level.
\begin{align*}
    \epsilon_{t}=\epsilon_0*e^{1-\frac{c_t}{\hat{c}}}
    \label{ls}
\end{align*}
where $\epsilon_0$ is the fixed value for vanilla label smoothing, $\hat{c}$ is the batch-level average confidence level.

\section{Experiments}

This section first exhibits empirical studies on the Quality Estimation (QE) task, a primary application of confidence estimation. Then, we present experimental results of our confidence-based label smoothing, an extension of our confidence estimate to better smoothing in NMT.

\subsection{Confidence-based Quality Estimation}

To evaluate the ability of our confidence estimate on mistake prediction, we experiment on extensive sentence/word-level QE tasks. Supervised QE task requires large amounts of parallel data annotated with the human evaluation, which is labor-intensive and impractical for low-resource languages. Here, we propose to address QE in an unsupervised way along with the training of the NMT model. 

\subsubsection{Sentence-level Quality Estimation}
We experiment on WMT2020 QE shared tasks\footnote{http://www.statmt.org/wmt20/quality-estimation-task.html}, including high-resource language pairs (English-German and English-Chinese) and mid-resource language pairs (Estonian-English and Romanian-English). This task provides source language sentences, corresponding machine translations, and NMT models used to generate translation. Each translation is annotated with direct assessment (DA) by professional translators, ranging from 0-100, according to the perceived translation quality. We can evaluate the performance of QE in terms of Pearson's correlation with DA scores.

We compare our confidence estimate with four unsupervised QE metrics \citep{DBLP:FomichevaSYBGFA20}:
\begin{itemize}
\item \textit{TP}: the sentence-level translation probability normalized by length $T$.
\item \textit{Softmax-Ent}: the average entropy of softmax output distribution at each decoding step.
\item \textit{Sent-Std}: the standard deviation of word-level log-probability $p(y_1),...,p(y_T)$.
\item \textit{D-TP}: the expectation for the set of TP scores by running $K$ stochastic forward passes through the NMT model with model parameters $\hat{\theta}^k$ perturbed by Monte Carlo (MC) dropout \citep{DBLP:conf/icml/GalG16}.
\end{itemize}



We also report two supervised QE models:
\begin{itemize}
    \item \textit{Predictor-Estimator} \citep{DBLP:conf/wmt/KimLN17}: a weak neural approach, which is usually set as the baseline system for supervised QE tasks. 
    \item \textit{BERT-BiRNN} \citep{DBLP:conf/wmt/KeplerTTVGFLM19}: a strong QE model using a large-scale dataset for pre-training and quality labels for fine-tuning.
\end{itemize}

We propose four confidence-based metrics: (1) \textit{Conf}: the sentence-level confidence estimate averaged by length, (2) \textit{Sent-Std-Conf}: the standard deviation of word-level log-confidence $c_1,...,c_T$, (3) \textit{D-Conf}: similar to D-TP, we compute the expectation of Conf by running $K$ forward passes through the NMT model, and (4) \textit{D-Comb}: the combination of D-TP and D-Conf:
\begin{align}
    {\rm D\mhyphen Comb}=\frac{1}{K}\sum^K_{k=1}({\rm Conf}_{\hat{\theta}^k}+{\rm TP}_{\hat{\theta}^k})
\end{align}

Note that our confidence estimate is produced together with translations. It is hard to let our model generate exact translations as provided by WMT, even with a similar configuration. Thus, we train our model on parallel sentences as used to train provided NMT models. Then, we employ force decoding on given translations to obtain existing unsupervised metrics and our estimations. We do not use any human judgment labels for supervision. 

Table~\ref{tab:QE1} shows the Pearson's correlation with DA scores for the above QE indicators. We find that:

\begin{table}[t]
\small
\centering
\begin{tabular}{lcccc}
    \toprule
    \multirow{2}{*}{Methods} & \multicolumn{2}{c}{Mid-resource} & \multicolumn{2}{c}{High-resource} \\ \cmidrule(lr){2-3} \cmidrule(lr){4-5}
     & Et-En & Ro-En & En-De & En-Zh \\ \midrule
    TP & 0.514 & 0.529 & 0.179 & 0.258 \\
    Softmax-Ent & 0.535 & 0.526 & 0.144 & 0.257 \\
    Sent-Std & 0.493 & 0.418 & 0.195 & 0.281 \\
    D-TP ($K$=30) & 0.583 & 0.553 & 0.197 & 0.288 \\ \midrule[0.5pt]
    Conf & 0.557 & 0.569 & 0.218 & \textbf{0.293} \\ 
    Sent-Std-Conf & 0.494 & 0.482 & \textbf{0.239} & 0.293 \\ 
    D-Conf ($K$=30) & 0.572 & 0.572 & 0.210 & 0.288 \\ 
    D-Comb ($K$=30) & \textbf{0.583} & \textbf{0.577} & 0.198 & 0.288 \\ \midrule
    PredEst \ddag & 0.477 & 0.685 & 0.145 & 0.190 \\ 
    BERT-BiRNN \ddag & 0.635 & 0.763 & 0.273 & 0.371 \\
    \bottomrule
\end{tabular}
\caption{Pearson's correlation between unsupervised QE indicators and DA scores. $K$ is set following \citet{DBLP:conf/emnlp/WangLWLS19}. We reimplement the first four unsupervised QE metrics on our NMT model. The best results of unsupervised metrics are marked in bold. Results with $\ddag$ are copies from \citet{DBLP:FomichevaSYBGFA20}.}
\label{tab:QE1}
\end{table}

Our confidence-based metrics substantially surpass probability-based metrics (the first three lines in Table~\ref{tab:QE1}). Compared with dropout-based methods (D-TP), our metrics obtain comparable results on mid-resource datasets while yielding better performance on high-resource translation tasks. We note that the benefits brought from the MC dropout strategy are limited for our metrics, which is significant in probability-based methods. It also proves the stability of our confidence estimate. In addition, the predictive power of MC dropout comes at the cost of computation, as performing forward passes through the NMT model is time-consuming and impractical for the large-scale dataset.

Our approach outperforms PredEst, a weak supervised method, on three tasks and further narrows the gap on Ro-En. Though existing unsupervised QE methods still fall behind with the strong QE model (BERT-BiRNN), the exploration of unsupervised metrics is also meaningful for real-world deployment with the limited annotated dataset.

\subsubsection{Word-level Quality Estimation}

\begin{table*}
\small
    \centering
    \begin{tabular}{lccccccccc}
    \toprule
    \multirow{2}{*}{Methods} & \multicolumn{6}{c}{Zh$\Rightarrow$En} & \multirow{2}{*}{En$\Rightarrow$De} & \multirow{2}{*}{De$\Rightarrow$En} & \multirow{2}{*}{Ro$\Rightarrow$En}\\ \cmidrule(lr){2-7}
     &  MT03 & MT04 & MT05 & MT06 & MT08 & ALL &  &  &  \\ \midrule[0.5pt]
    Transformer w/o LS &  48.77 & 48.50 & 47.45 & 46.65 & 35.93 & 45.50 & 26.98 & 34.27 & 29.71 \\
    + Standard LS &  49.14 & 48.48 & 50.53 & 47.44 & 36.23 & 45.83 & 27.40 & 34.52 & 30.03 \\
    + Confidence-based LS  &  $50.2^*$ & 48.57 & $50.91^*$ & $48.57^*$ & $37.38^*$ & $46.55^*$ & $27.75^*$ & $35.02^*$ & $30.82^*$ \\\bottomrule
    \end{tabular}
    \caption{Translation results (beam size 4) for standard label smoothing and our confidence-based label smoothing on NIST Zh$\Rightarrow$En, WMT14 En$\Rightarrow$De (using case-sensitive BLEU score for evaluation), IWSLT14 De$\Rightarrow$En, and WMT16 Ro$\Rightarrow$En. ``$\ast$'' indicates gains are statistically significant than Transformer w/o LS with $p<0.05$.}
    \label{tab:my_label}
\end{table*}

\begin{table}
    \centering
    \small
    \begin{tabular}{l|ccccc}
    \toprule
    Methods & AUROC$\uparrow$ & AUPR$\uparrow$ & EER$\downarrow$ & DET$\downarrow$ \\ \midrule
    MSP & 72.59 & 97.49 & 32.30 &  31.22 \\
    MCDropout & 86.52 & 99.23 & 20.80 & 20.76\\
    Ours & 85.89 & 99.07 & 20.40 &  19.90 \\
    \bottomrule
    \end{tabular}
    \caption{Word-level QE evaluated by the separation accuracy of OK and BAD translations in the Zh$\Rightarrow$En task. All values are shown in percentages. $\uparrow$ indicates higher scores are better, and $\downarrow$ indicates lower is better.}
    \label{tab:QE2}
\end{table}

We also validate the effectiveness of our confidence estimate on QE tasks from a more fine-grained view. We randomly select 250 sentences from Zh$\Rightarrow$En NIST03 and obtain NMT translations. Two graduate students are asked to annotate each target word as either OK or BAD. We assess the performance of failure prediction with standard metrics, which are introduced in Appendix~\ref{sec:qe}.

Experimental results are given in Table~\ref{tab:QE2}. We implement competitive failure prediction approaches, including Maximum Softmax Probability (MSP) \citep{DBLP:conf/iclr/HendrycksG17} and Monte Carlo Dropout (MCDropout) \citep{DBLP:conf/icml/GalG16}. We find that our learned confidence estimate yields a better separation of OK and BAD translation than MSP. Compared with MCDropout, our metrics achieve competing performance with significant advantages on computational expenses. 


Overall, the learned confidence estimate is a competitive indicator of translation precision compared with other unsupervised QE metrics. Moreover, the confidence branch added to the NMT system is a light component. It allows each translation to come with quality measurement without degradation of the translation accuracy. The performance with the confidence branch is in Appendix \ref{sec:BLEU_with_conNet}.

\subsection{Confidence-based Label Smoothing}

We extend our confidence estimate to improve smoothing and experiment on different-scale translation tasks: WMT14 English-to-German (En$\Rightarrow$De), LDC Chinese-to-English (Zh$\Rightarrow$En)\footnote{The corpora includes LDC2000T50, LDC2002T01, LDC2002E18, LDC2003E07, LDC2003E14, LDC2003T17, and LDC2004T07.}, WMT16 Romanian-to-English (Ro$\Rightarrow$En), and IWSLT14 German-to-English (De$\Rightarrow$En). We use the 4-gram BLEU \cite{papineni2002bleu:} to score the performance. More details about data processing and experimental settings are in Appendix~\ref{sec:hyper}.

As shown in Table~\ref{tab:my_label}, our confidence-based label smoothing outperforms standard label smoothing by adaptively tuning the amount of each label smoothing. For Zh$\Rightarrow$En task, our method improves the performance over Transformer w/o LS by 1.05 BLEU, which also exceeds standard label smoothing by 0.72 BLEU. We find that improvements over standard label smoothing differ in other language pairs (0.35 BLEU in En$\Rightarrow$De, 0.5 BLEU in De$\Rightarrow$En, and 0.79 BLEU in Ro$\Rightarrow$En). It can be attributed to that the seriousness of miscalibration varies in different language pairs and datasets \citep{DBLP:conf/acl/WangTSL20}.

Experimental results with a larger search space (i.e. beam size=30) are also given in Appendix~\ref{sec:hyper} to support the above findings. 

\section{Analysis}

Confidence estimation is particularly critical in real-world deployment, where noisy samples and out-of-distribution data are prevalent \citep{DBLP:conf/nips/SnoekOFLNSDRN19}. Given those abnormal inputs, neural network models are prone to be highly confident in misclassification \citep{DBLP:conf/cvpr/NguyenYC15}. Thus, we need an accurate confidence estimate to detect potential failures caused by odd inputs by assigning them low confidence. This section explores whether our confidence estimate can accurately measure risk under those two conditions.

\subsection{Noisy Label Identification}

\begin{table*}[]
\small
    \centering
    \begin{tabular}{ccccc}
    \toprule
    \multirow{2}{*}{Noise Rate} & AUROC$\uparrow$ & AUPR$\uparrow$ & EER$\downarrow$ & DET$\downarrow$ \\ \cmidrule(lr){2-5}
    & \multicolumn{4}{c}{The Model Probability / Our Confidence Estimate} \\ \midrule[0.5pt]
    20\% & 93.21 / \textbf{96.73} & 97.08 / \textbf{98.57} & 13.50 / \textbf{7.00} & 11.50 / \textbf{6.00} \\
    40\% & 94.89 / \textbf{95.73} & 95.22 / \textbf{95.50} & 11.88 / \textbf{9.50} & 10.58 / \textbf{7.69} \\
    60\% & 93.37 / \textbf{94.92} & 86.54 / \textbf{88.09} & 14.00 / \textbf{10.08} & 12.04 / \textbf{8.29} \\
    80\% &  91.63 / \textbf{95.44} & 64.15 / \textbf{76.67} & 16.06 / \textbf{10.13} & 13.41 / \textbf{8.13} \\ \bottomrule
    \end{tabular}
    \caption{Separating clean and noisy data by the model probability and our confidence estimate with varying noisy rates. $\uparrow$ indicates that higher scores are better, while $\downarrow$ means that lower is better. All values are percentages.}
    \label{tab:noise}
\end{table*}

\begin{table*}
\small
    \centering
    \begin{tabular}{lcccccc}
    \toprule
    \multicolumn{3}{c}{Out-of-distribution Dataset} & AUROC$\uparrow$ & AUPR$\uparrow$ & EER$\downarrow$ & DET$\downarrow$ \\ \cmidrule(lr){1-3} \cmidrule(lr){4-7}
    Corpus & UNK & Len. & \multicolumn{4}{c}{The Model Probability / Our Confidence Estimate} \\ \midrule[0.5pt]
    WMT-News & 1.45\% & 30.16 & 71.51 / \textbf{72.01} & 68.86 / \textbf{70.97} & \textbf{33.78} / 34.44 & 33.33 / \textbf{32.44} \\
    Tanzil & 1.36\% & 34.17  & \textbf{90.53} / 89.48 & \textbf{91.45} / 91.32 & \textbf{17.33} / 18.78 & \textbf{16.72} / 17.72 \\
    Tico-19 & 1.21\% & 30.29 & 64.10 / \textbf{72.10} & 62.12 / \textbf{71.59} & 39.67 / \textbf{33.33} & 38.83 / \textbf{31.83}\\      
    TED2013 & 1.04\% & 19.03 & 63.48 / \textbf{68.44} & 59.10 / \textbf{66.75} & 39.22 / \textbf{36.22} & 39.00 / \textbf{35.39} \\
    News-Commentary & 1.00\% & 23.81 & 64.14 / \textbf{70.10} & 60.49 / \textbf{69.48} & 39.33 / \textbf{35.56} & 39.11 / \textbf{34.22} \\ \bottomrule
    \end{tabular}
    \caption{Comparison of the model probability and our confidence estimate on out-of-domain data detection tasks. We present the rate of unknown words (UNK) and average length of input sentences for each dataset (the average input length of in-domain dataset is 22.47). All scores are shown in percentages and the best results are highlighted in bold. $\uparrow$ indicates that higher scores are better, while $\downarrow$ indicates that lower scores are better.}
    \label{tab:ood}
\end{table*}

\begin{figure}
	\centering
	\includegraphics[width=.43\textwidth]{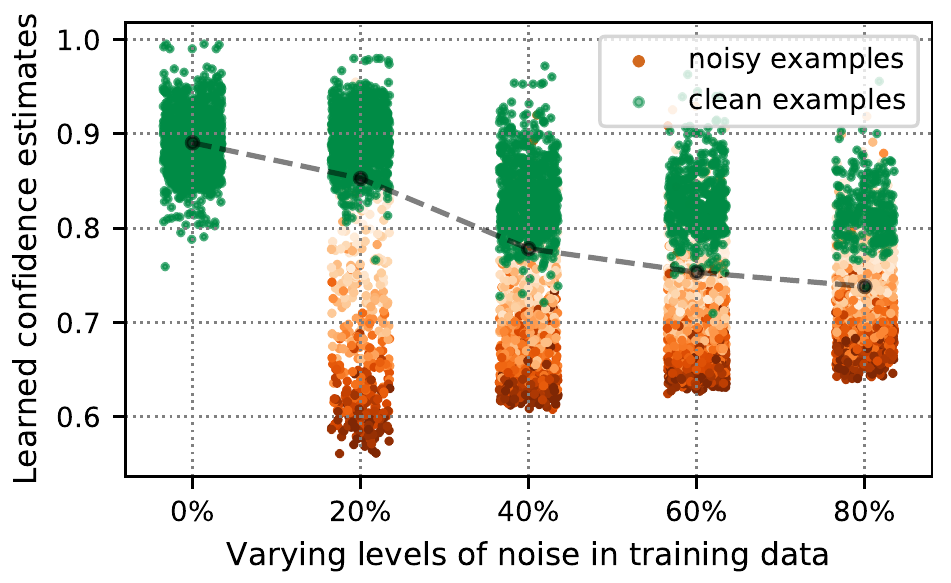}
	\caption{The learned confidence estimate on IWSLT14 De$\Rightarrow$En as varying levels of noisy labels. The shade of colors denotes how many words are corrupted in a sentence (dark orange means a high pollution rate). The dashed line shows averaged learned confidence estimate on the whole dataset.}
	\label{fig:visual}
\end{figure}

We expect that the model requires more hints to fit noisy labels by predicting low confidence. To test this point, we experiment on the IWSLT14 De$\Rightarrow$En dataset containing 160k parallel sentences. We build several datasets with progressively increasing noisy samples by randomly replacing target-side words with others in the vocabulary. We train on each dataset with the same configuration and picture the learned confidence estimate in Figure \ref{fig:visual}.

The learned confidence estimate appears to make reasonable assessments. (1) It predicts low confidence on noisy samples but high confidence on clean ones. Specifically, the confidence estimate is much lower as a higher pollution degree in one example (darker in color). (2) With increasing noises in the dataset, the NMT model becomes more uncertain about its decision accordingly. Large numbers of noises also raise a challenge for separating clean and noisy samples. 

We also compare ours with the model probability by giving the accuracy of separating clean and noisy examples under varying pollution rates. We set clean data as the positive example and use evaluation metrics listed in Appendix \ref{sec:qe}.

As shown in Table \ref{tab:noise}, our confidence estimate obtains better results in all cases, especially in a high noise rate. Our metric improves the area under the precision-recall curve (AUPR) from 64.15\% to 76.76\% and reduces the detection error (DET) from 13.41\% to 8.13\% at an 80\% noise rate. It proves that our confidence estimate is more reliable for detecting potential risks induced by noisy data.


\subsection{Out-of-Domain Data Detection}


For our in-domain examples, we train an NMT model on the 2.1M LDC Zh$\Rightarrow$En news dataset and then sample 1k sentences from NIST2004 as the in-domain testbed. We select five out-of-domain datasets and extract 1k samples from each. Most of them are available for download on OPUS, specified in Appendix \ref{sec:ood}. Regarding the unknown words (UNK) rate, the average length of input sentences, and domain diversity, the descending order based on distance with the in-domain dataset is WMT-news > Tanzil > Tico-19 > TED2013 > News-Commentary. Test sets closer to the in-domain dataset are intuitively harder to tell apart. 

We use sentence-level posterior probability and confidence estimate of the translation to separate in- and out-of-domain data. Evaluation metrics are in Appendix~\ref{sec:qe}. Results are given in Table \ref{tab:ood}.


We find that our approach performs comparably with the probability-based method on datasets with distinct domains (WMT-news and Tanzil). But when cross-domain knowledge is harder to detect (the last three lines in Table \ref{tab:ood}), our metric yields a better separation of in- and out-of-domain ones. 

\begin{figure}[t]
	\centering
	\subfigure[Rank by prediction probability] {\includegraphics[width=.48\textwidth]{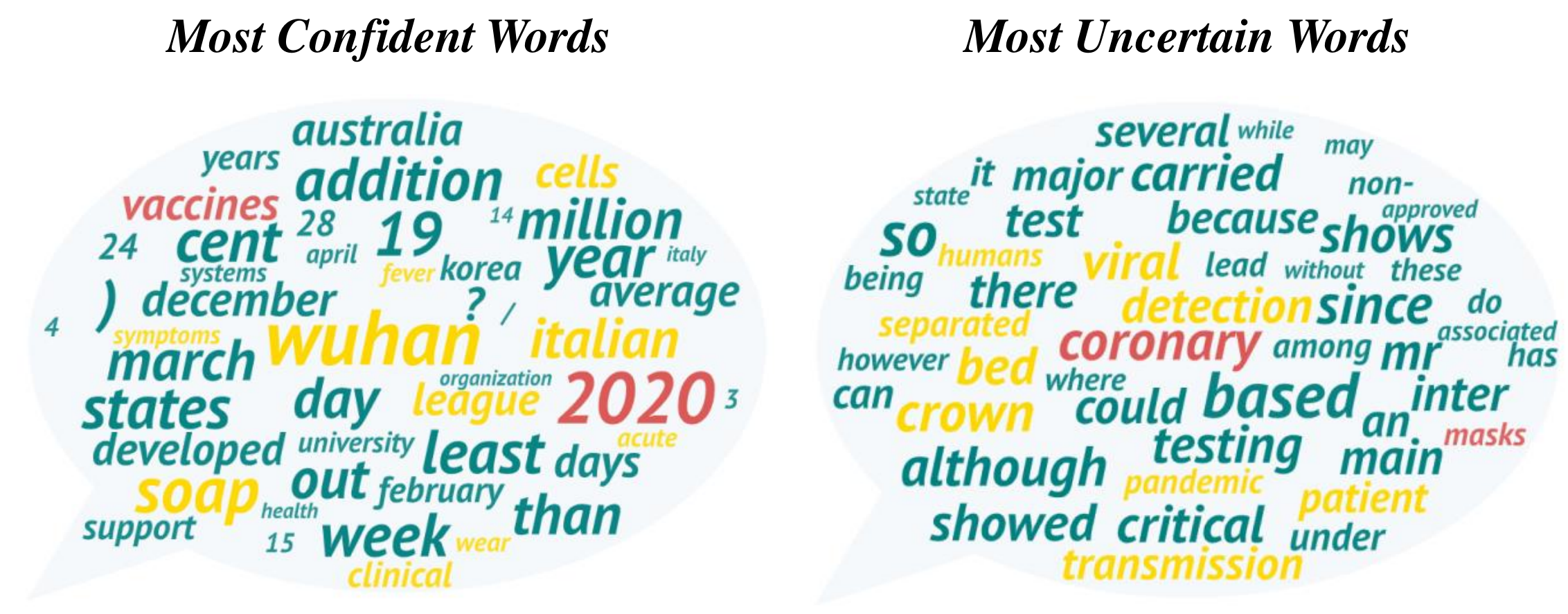}
	\label{fig:word_cloud_p}} 
	\\
	\subfigure[Rank by our confidence estimate] {\includegraphics[width=.48\textwidth]{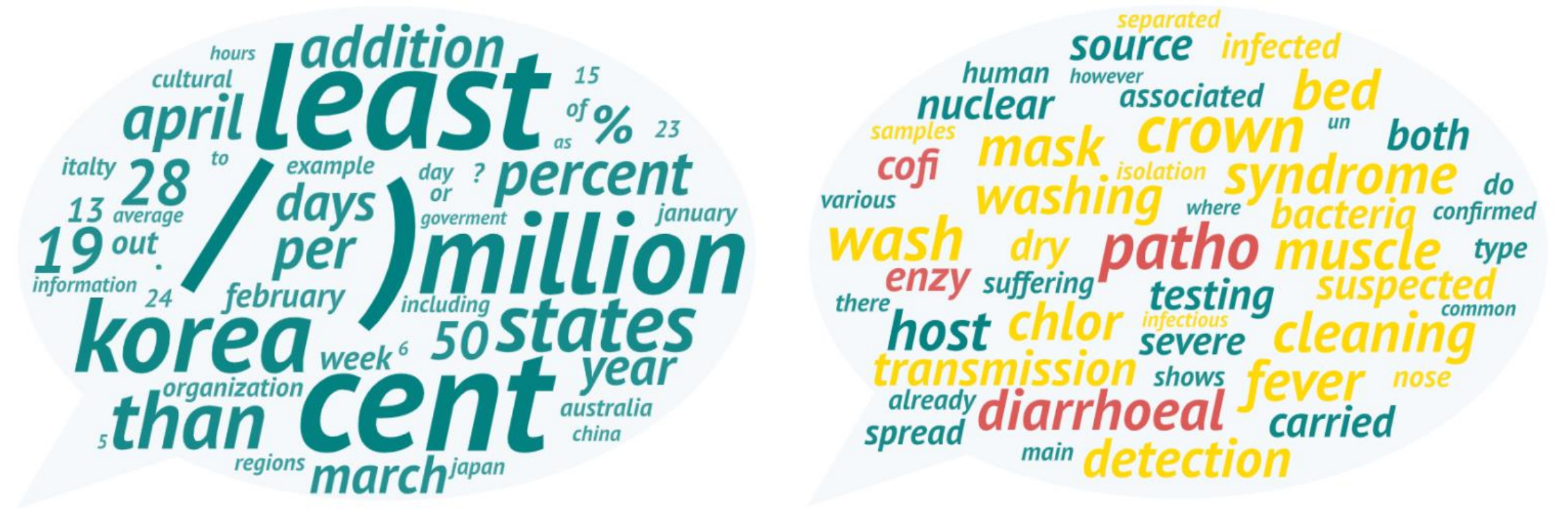}
	\label{fig:word_cloud_c}}
	\caption{Word clouds of the most confident/uncertain translations in the Tico-19 dataset ranked by (a) prediction probability and (b) learned confidence estimate. We divide tokens into three categories based on their frequencies.  \textcolor[rgb]{0.179,0.542,0.3398}{\textit{High}}: the most 3k frequent words, \textcolor[rgb]{1, 0.71, 0.06}{\textit{Medium}}: the most 3k-12k frequent words, \textcolor[rgb]{0.855,0.343,0.336}{\textit{Low}}: the other tokens.}
	\label{fig:word_cloud}
\end{figure}

To better understand the behaviour of our confidence estimates on out-of-domain data, we visualize word clouds of the most confident/uncertain words ranked by model probability and our measurements on a medicine dataset (Tico-19) in Figure~\ref{fig:word_cloud}. The colors of words indicate their frequencies in the in-domain dataset. 


Our metrics correctly separate in- and out-of-domain data from two aspects: (1) \textit{word frequency}: the NMT model is certain about frequent words yet hesitates on rare words as seen in Figure~\ref{fig:word_cloud_c}. But colors in Figure~\ref{fig:word_cloud_p} are relatively mixing. (2) \textit{domain relation}: the most uncertain words ranked by our confidence estimate are domain-related, like ``patho'' and ``syndrome'', while the most confident words are domain-unrelated (e.g., punctuations and prepositions). This phenomenon cannot be seen in Figure~\ref{fig:word_cloud_p}, showing that probabilities from softmax fall short in representing model uncertainty for domain-shift data.

\section{Related Work}

The task of confidence estimation is crucial in real-world conditions, which helps failure prediction \citep{DBLP:conf/nips/CorbiereTBCP19} and out-of-distribution detection \citep{DBLP:conf/iclr/HendrycksG17,DBLP:conf/nips/SnoekOFLNSDRN19,lee2018a}. This section reviews recent researches on confidence estimation and related applications on quality estimation for NMT.

\subsection{Confidence Estimation for NMT}


Only a few studies have investigated calibration in NMT. \citet{DBLP:conf/nips/MullerKH19} find that the NMT model is well-calibrated in training, which is proven severely miscalibrated in inference \citep{DBLP:conf/acl/WangTSL20}, especially when predicting the end of a sentence \citep{DBLP:journals/corr/abs-1903-00802}. Regarding the complex structures of NMT, the exploration for fixing miscalibration in NMT is scarce. \citet{DBLP:conf/emnlp/WangLWLS19, DBLP:journals/corr/abs-2006-08344} use Monte Carlo dropout to capture uncertainty in NMT, which is time-consuming and computationally expensive. Unlike them, we are the first to introduce learned confidence estimate into NMT. Our method is well-designed to adapt to Transformer architecture and NMT tasks, which is also simple but effective.

\subsection{Quality Estimation for NMT}

QE is to predict the quality of the translation provided by an MT system at test time without standard references. Recent supervised QE models are resource-heavy and require a large mass of annotated quality labels for training \citep{wang-etal-2018-alibaba,kepler-etal-2019-unbabels,lu2020quality}, which is labor-consuming and unavailable for low-resource languages. 

Exploring internal information from the NMT system to indicate translation quality is another alternative. \citet{DBLP:FomichevaSYBGFA20} find that uncertainty quantification is competitive in predicting the translation quality, which is also complementary to supervised QE model \citep{wang-etal-2021-beyond-glass}. However, they rely on repeated Monte Carlo dropout \citep{DBLP:conf/icml/GalG16} to assess uncertainty at the high cost of computation. Our confidence estimate outperforms existing unsupervised QE metrics, which is also intuitive and easy to implement. 

\section{Conclusion}

In this paper, we propose to learn confidence estimates for NMT jointly with the training process. We demonstrate that learned confidence can better indicate translation accuracy on extensive sentence/word-level QE tasks and precisely measures potential risk induced by noisy samples or out-of-domain data. We further extend the learned confidence estimate to improve smoothing, outperforming the standard label smoothing technique. As our confidence estimate outlines how much the model knows, we plan to apply our work to design a more suitable curriculum during training and post-edit low-confidence translations in the future.

\section*{Acknowledgements}
This work is supported by the Natural Science Foundation of China under Grant No. 62122088, U1836221, and 62006224.



\bibliography{anthology}
\bibliographystyle{acl_natbib}

\clearpage
\appendix

\section{Evaluation Metrics}
\label{sec:qe}
We let TP, FP, TN, and FN represent true positives, false positives, true negatives, and false negatives. We use the following metrics for evaluating the accuracy of word-level QE, noisy label identification, and out-of-domain detection:

\begin{itemize}
    \item \textit{AUROC}: the Area Under the Receiver Operating Characteristic (ROC) curve, which plots the relation between TPR and FPR. 
    \item \textit{AUPR}: the Area Under the Precision-Recall (PR) curve. The PR curve is made by plotting \textit{precision} = TP/(TP+FP) and \textit{recall} = TP/(TP+FN).
    \item \textit{DET}: the Detection Error, which is the minimum possible misclassification probability over all possible threshold when separating positive and negative examples.
    \item \textit{EER}: the Equal error rate. It is the error rate when the confidence threshold is located where FPR is the same with the false negative rate (FNR) = FN / (TP+FN).
\end{itemize}

We set OK translations in the word-level QE task, clean samples in the noisy data identification task, and in-domain samples in the out-of-domain data detection task as the positive example.

\section{Translation Results with the Confidence Branch}
\label{sec:BLEU_with_conNet}

\begin{table*}
\small
    \centering
    \begin{tabular}{lcccccccc}
    \toprule
    \multirow{2}{*}{Methods} & \multicolumn{6}{c}{Zh$\Rightarrow$En} & \multirow{2}{*}{En$\Rightarrow$De} & \multirow{2}{*}{De$\Rightarrow$En} \\ \cmidrule(lr){2-7}
     &  MT03 & MT04 & MT05 & MT06 & MT08 & ALL & \\ \midrule[0.5pt]
    Transformer &  49.14 & 48.48 & 50.53 & 47.44 & 36.23 & 45.83 & 27.40 & 34.52\\
    + ConNet &  49.51 & 48.47 & 50.51 & 47.29 & 36.44 & 45.90 & 27.55 & 34.73\\
    \bottomrule
    \end{tabular}
    \caption{Translation results (BLEU score) with the confidence branch on NIST Zh$\Rightarrow$En, WMT14 En$\Rightarrow$De (using case-sensitive BLEU score for evaluation) and IWSLT14 De$\Rightarrow$En.}\label{tab:with_conNet}
\end{table*}

The confidence branch added to the NMT system is a light component. It allows each translation to come with quality measurement without degradation of the translation accuracy. Translation results with the confidence branch are given in Table \ref{tab:with_conNet}. 

We see that the added confidence branch does not affect the translation performance. Implementation details in section \ref{method} are necessary for achieving this. For instance, if we use the highest hidden state to predict confidence and translation together, BLEU scores would dramatically decline with a larger beam size, the drop of which is more significant than that of the baseline model. For the En$\Rightarrow$De task, the change is from 27.31 (beam size 4) to 25.6 (beam size 100), while the baseline model even improves 0.5 BLEU further with a larger beam size 100.

\section{Confidence-based Label Smoothing}
\label{sec:hyper}

We experiment on different-scale translation tasks: WMT14 En$\Rightarrow$De, LDC Zh$\Rightarrow$En, WMT16 Ro$\Rightarrow$En, and IWSLT14 De$\Rightarrow$En.

\paragraph{Datasets.} We tokenize the corpora by Moses \citep{DBLP:conf/acl/KoehnHBCFBCSMZDBCH07}. Byte pair encoding (BPE) \citep{sennrich2015neural} is applied to all language pairs to construct a join 32k vocabulary except for Zh$\Rightarrow$En where the source and target languages are separately encoded.

For En$\Rightarrow$De, we train on 4.5M training samples. Newstest2013 and newstest2014 are set as validation and test sets. For Zh$\Rightarrow$En, we remove sentences of more than 50 words and collect 2.1M training samples. We use NIST 2002 as the validation set, NIST 2003-2006 (MT03-06), and 2008 (MT08) as the testbed. For Ro$\Rightarrow$En, we train on 0.61M training data and use newsdev2016 and newstest2016 as validation and test sets. For De$\Rightarrow$En, we train on its training set with 160k training samples and evaluate on its test set. 

\paragraph{Settings.} We implement the described model with fairseq\footnote[5]{https://github.com/pytorch/fairseq} toolkit for training and evaluating. We follow \citet{vaswani2017attention} to set the configurations of models with the \texttt{base} Transformer. The dropout rate of the residual connection is 0.1 except for Zh$\Rightarrow$En (0.3). The experiments last for 150k steps for Zh$\Rightarrow$En and En$\Rightarrow$De, 30k for small-scale De$\Rightarrow$En and Ro$\Rightarrow$En. We average the last ten checkpoints for evaluation and adopt beam search (beam size 4/30, length penalty 0.6). We set $\epsilon_{ls}=0.1$ for the vanilla label smoothing. 


The hyper-parameters $\lambda_0$ and $\beta_0$ (as seen Equation~\ref{anneal}) control the initial value and declining speed of $\lambda$ (as in Equation~\ref{loss}), which decides the number of hints the NMT model can receive. To ensure that no hints are available at the early stage of training, we set $\lambda_0=30$, $\beta_0=4.5*10^4$ for Zh$\Rightarrow$En and En$\Rightarrow$De, $\beta_0=1.2*10^4$ for De$\Rightarrow$En and Ro$\Rightarrow$En. We set $\epsilon_0=0.1$ (as seen in Equation~\ref{ls}) for all language pairs.

\begin{table*}
\small
    \centering
    \begin{tabular}{lccccccccc}
    \toprule
    \multirow{2}{*}{Methods} & \multicolumn{6}{c}{Zh$\Rightarrow$En} & \multirow{2}{*}{En$\Rightarrow$De} & \multirow{2}{*}{De$\Rightarrow$En} & \multirow{2}{*}{Ro$\Rightarrow$En}\\ \cmidrule(lr){2-7}
     &  MT03 & MT04 & MT05 & MT06 & MT08 & ALL &  &  &  \\ \midrule[0.5pt]
    Transformer w/o LS &  49.06 & 48.64 & 47.76 & 47.01 & 35.93 & 45.68 & 25.91 & 34.36 & 29.96 \\
    + Standard LS &  49.63 & 48.70 & 50.61 & 47.81 & 37.61 & 46.27 & 27.81 & 34.66 & 30.48 \\
    + Confidence-based LS  &  $50.59^*$ & 48.75 & $51.47^*$ & $48.60^*$ & $37.87^*$ & $46.85^*$ & $28.01^*$ & $35.11^*$ & $31.07^*$ \\\bottomrule
    \end{tabular}
    \caption{Translation results (beam size 30) for standard label smoothing and our confidence-based label smoothing on NIST Zh$\Rightarrow$En, WMT14 En$\Rightarrow$De (using case-sensitive BLEU score for evaluation), IWSLT14 De$\Rightarrow$En, and WMT16 Ro$\Rightarrow$En. ``$\ast$'' indicates gains are statistically significant than Transformer w/o LS with $p<0.05$.}
    \label{tab:my_label_30}
\end{table*}

\paragraph{Results.} A common setting with beam size=4 is given in Table~\ref{tab:my_label} in the main body. Here, we experiment with a larger search space where being over-or under-confident further worsens model performance~\citep{DBLP:conf/icml/GuoPSW17}. The results with beam size=30 are listed in Table~\ref{tab:my_label_30}. For Zh$\Rightarrow$En task, our method yields +1.17 BLEU improvements over Transformer w/o LS, exceeding standard label smoothing by 0.58 BLEU scores. The performance gains can also be found in other language pairs, showing the effectiveness of our confidence-based label smoothing with a larger beam size.




\section{Out-of-domain Data Detection}

We select five out-of-domain datasets for our tests (we extract 1k samples each), which are available for download on OPUS\footnote{https://opus.nlpl.eu/}. The datasets are:

\label{sec:ood}
\begin{itemize}
    \item \textit{WMT-News}: A parallel corpus of News Test Sets provided by WMT for training SMT\footnote{http://www.statmt.org/wmt19/}, which is rich in content including sports, entertainment, politics, and so on.
    \item \textit{Tanzil}: This is a collection of Quran translations compiled by the Tanzil project\footnote{https://opus.nlpl.eu/Tanzil-v1.php}.
    \item \textit{Tico-19}: This is a collection of translation memories from the Translation Initiative for COVID-19, which has many medical terms\footnote{https://opus.nlpl.eu/tico-19-v2020-10-28.php}.
    \item \textit{TED2013}: A corpus of TED talks subtitles provided by CASMACAT\footnote{http://www.casmacat.eu/corpus/ted2013.html}, which are about personal experiences in informal expression.
    \item \textit{News-Commentary}: It is also a dataset provided by WMT\footnote{https://opus.nlpl.eu/News-Commentary-v16.php}, but the extracted test set is all about international politics.
\end{itemize}

\end{document}